\documentclass[10pt,journal,ruled,linesnumbered]{IEEEtran}
\usepackage[latin9]{inputenc}
\usepackage{color}
\usepackage{booktabs}
\usepackage{multirow}
\usepackage{algorithm2e}
\usepackage{amsmath}
\usepackage{amssymb}
\usepackage{graphicx}

\makeatletter

\usepackage{epsfig}
\usepackage[compress]{cite}
\usepackage{times}
\allowdisplaybreaks
\makeatother

\begin{document}
	\title{CT-VIR: Continuous-Time Visual-Inertial-Ranging Fusion for Indoor Localization with Sparse Anchors\thanks{This work was supported by the National Natural Science Foundation
			of China (61972131).}}
	\author{~Yu-An~Liu,~Li~Zhang
		\thanks{Y. Liu and L. Zhang are with the School of Mathematics, Hefei University
			of Technology, Hefei 230031, China (e-mail: yuanliu0015@gmail.com
			(Y. Liu); lizhang@hfut.edu.cn (L.Zhang)}\thanks{Manuscript received ; revised .}}
	\maketitle
	\begin{abstract}
		\textcolor{black}{Visual-inertial odometry (VIO) is widely used for mobile robot localization, but its long-term accuracy degrades without global constraints. Incorporating ranging sensors such as ultra-wideband (UWB) can mitigate drift, yet high-accuracy ranging usually requires well-deployed anchors, which is hard to ensure in narrow or low-power environments. Moreover, most existing visual-inertial-ranging (VIR) fusion methods rely on discrete time-based filtering or optimization, making it difficult to balance positioning accuracy, trajectory consistency, and fusion efficiency under asynchronous multi-sensor sampling. To address these issues, we propose a spline-based continuous-time state estimation method for VIR fusion localization. In the preprocessing stage, VIO motion priors and UWB ranging measurements are used to construct virtual anchors and to reject outliers, alleviating geometric degeneration and improving range reliability. In the estimation stage, the pose trajectory is parameterized in continuous time using a B-spline, while inertial, visual, and ranging constraints are formulated as factors in a sliding-window graph and the spline control points together with a small set of auxiliary parameters are jointly optimized to obtain a continuous-time trajectory estimate. Evaluations on public datasets and real-world experiments demonstrate the effectiveness and practical potential of the proposed approach.}
	\end{abstract}
	

\section{Introduction}

Mobile robots, such as unmanned aerial vehicles (UAVs), have become key platforms for a wide range of autonomous applications in complex environments \cite{1,2,3}. Consequently, accurate and robust state estimation is a fundamental requirement for their reliable operation \cite{4,5,6}. Among existing techniques, visual-inertial odometry (VIO) has emerged as a widely used pose estimation method due to its low cost and high short-term accuracy \cite{7,8,9}, achieved by tightly fusing camera and IMU measurements without relying on external infrastructure. However, in the absence of global constraints, VIO inevitably suffers from long-term drift and degraded performance in texture-poor, motion-blurred, or partially occluded scenes \cite{10}.

To reduce drift and improve global consistency, many studies incorporate multi-sensor fusion into VIO pipelines, such as LiDAR \cite{11}, millimeter-wave radar \cite{12}, or wheel odometry \cite{13}. It is worth mentioning that adding range measurements is particularly attractive, as they provide absolute distance information to the environment or external references, bounding cumulative drift and enhancing robustness under visually degraded or complex conditions \cite{14,15,16}. In open outdoor environments, global navigation satellite system (GNSS) is typically used \cite{17,17.1}, whereas in GNSS-denied scenarios UWB-based ranging data offers a practical alternative \cite{18,18.1}. However, high-accuracy ranging critically depends on the number and geometry of physical anchors. Deploying and maintaining a sufficient number of well-calibrated anchors with good spatial coverage is challenging in narrow indoor spaces, underground facilities, or low-power environments \cite{19}. Besides, UWB ranging is vulnerable to non-line-of-sight (NLOS) propagation and multipath, which introduce biased, heavy-tailed errors that can severely degrade performance if not handled robustly \cite{20}.

Another challenge arises from the multi-rate and asynchronous nature of visual-inertial-ranging (VIR) data. Most existing multi-sensor fusion is formulated in a discrete-time framework, typically following either a recursive filter or a batch optimization paradigm: the former relies on the Kalman filter family and its extensions for online recursive state estimation \cite{21,22,23}, while the latter performs nonlinear optimal estimation by jointly optimizing states within a temporal window \cite{24,25,26}. Recent work such as an EKF-based VIR fusion framework is proposed in \cite{27}, and it was augmented in \cite{28} with noise statistical modeling and adaptive estimation to handle degraded measurements, and a hierarchical hybrid strategy was introduced that combines filtering and optimization in a layered manner \cite{29}. Although effective in many scenarios, these approaches propagate the state between discrete nodes using simplified motion models and approximate measurement timestamps by snapping them to the nearest keyframe or IMU integration interval. As a result, it is difficult to simultaneously guarantee high localization accuracy, temporal consistency, and computational efficiency when sensor rates differ significantly or ranging updates are irregular and sparse. Besides, in anchor-scarce settings, additional interpolation at misaligned timestamps can further amplify range noise and outliers, making the system more sensitive to NLOS and multipath effects.

Beyond discrete-time formulations, continuous-time representations have been increasingly adopted in the literature for modeling complex dynamical systems \cite{30,31,32}. For UAV trajectory estimation tasks that demand high trajectory smoothness, temporal consistency, and precise alignment of asynchronous multi-sensor data, continuous-time trajectory representations offer an appealing alternative: by modeling the robot motion as a smooth spline curve in $SE(3)$ \cite{33}, heterogeneous measurements can be associated with their exact timestamps, and analytic trajectory derivatives can be used to incorporate inertial constraints in a principled manner \cite{34}. Such continuous-time formulations have been successfully applied in multi-sensor fusion systems and shown clear advantages in handling highly asynchronous sensors \cite{35}, as evidenced by recent works including related LiDAR-inertial-camera fusion methods \cite{36,37,38}, continuous-time UWB-inertial fusion that exploits sparse anchors and irregular ranging \cite{39}, Ctrl-VIO for rolling-shutter VIO \cite{40}, and CT-UIO \cite{41}, which targets few-anchor UWB-inertial-odometer localization using non-uniform B-splines.

To the best of our knowledge, a unified continuous-time framework that tightly fuses visual, inertial, and UWB ranging under anchor-scarce, NLOS-degraded conditions remains largely unexplored. In particular, how to exploit continuous-time modeling to alleviate geometric degeneration and error amplification induced by sparse, asynchronous UWB measurements is still an open problem, whereas conventional discrete-time formulations often exacerbate these issues through interpolation and re-integration. To bridge this gap, we propose a spline-based VIR fusion framework that leverages VIO-derived motion priors to robustly reject inconsistent UWB outliers and to construct local virtual anchors from short motion-range segments, so that both physical and virtual anchors jointly provide geometrically informative constraints to a continuous-time B-spline trajectory estimated in a factor-graph back-end.

The main contributions of this paper are as follows: (i) We propose a spline-based continuous-time VIR fusion framework that models the pose trajectory with a cubic B-spline and encodes visual, inertial, and ranging measurements as factors in a sliding-window graph, jointly targeting high localization accuracy, temporal consistency, and computational efficiency under asynchronous sampling. (ii) We develop a motion-prior-guided ranging preprocessing and virtual-anchor construction scheme that combines robust median/MAD-based outlier rejection with local least-squares fitting and information/geometry checks, effectively alleviating geometric degeneration and enhancing observability in few-anchor or partially failed-anchor settings. (iii) We carry out extensive experiments on multiple public datasets and real-world flights with different anchor configurations, showing that the proposed method achieves accurate and robust localization compared with representative continuous-time and discrete-time fusion baselines.

The remainder of this paper is organized as follows. Section 2 gives an overview of the proposed framework. Section 3 formulates the continuous-time VIR localization problem. Section 4 presents the methodology, including visual-inertial preprocessing, ranging preprocessing with virtual anchors, and the continuous-time B-spline factor-graph. Section 5 reports experimental results on public datasets and real-world scenarios, and Section 6 concludes the paper and outlines future work.

\section{System Overview }

\begin{figure*}[th]
	\centering \includegraphics[scale=0.24]{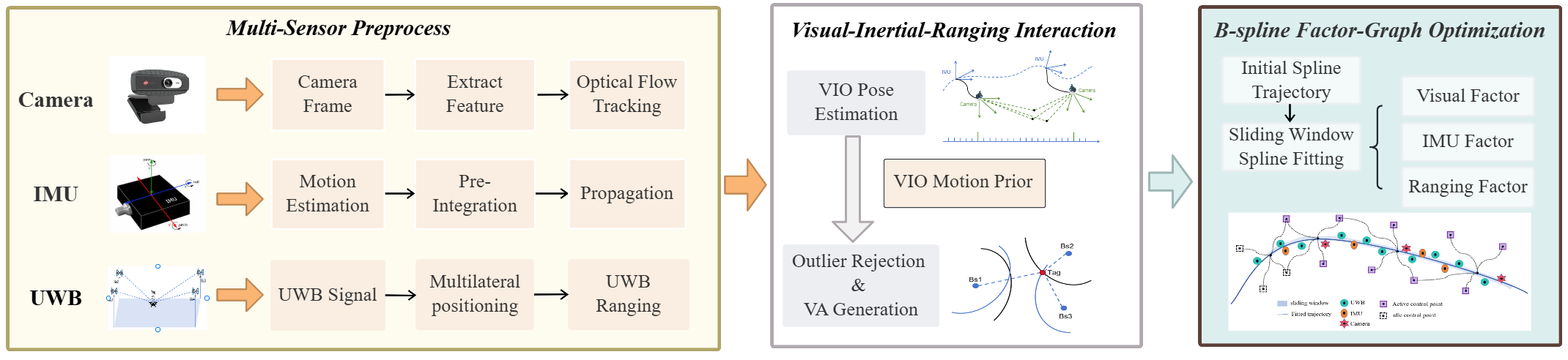}
	\protect\caption{System overview of the proposed continuous-time VIR localization framework.}
	\label{figure:1}
\end{figure*}

As illustrated in Figure 1, the proposed system is a spline-based continuous-time VIR localization framework. A camera, IMU, and UWB transceivers are rigidly mounted on the robot platform: the camera acquires images for feature extraction and tracking, the IMU provides high-rate motion measurements, and the UWB modules communicate with a small number of static anchors to obtain one-way range observations. All sensors are time-synchronized and extrinsically calibrated to a common reference frame. Next, the raw observations are converted into higher-level motion and ranging constraints. A tightly coupled VIO front-end takes camera and IMU data as input and outputs short-horizon pose estimates as motion priors. These priors are used both to initialize the continuous-time trajectory and to predict the robot position at each timestamp for checking the consistency of UWB measurements (initialized by VIO and then updated by the current spline estimate from the previous sliding window), based on which a robust range outlier rejection procedure is applied. In the back-end, we parameterize the robot pose trajectory in continuous time using a cubic B-spline with uniformly spaced knots, and model prior, IMU, visual reprojection, and physical/virtual anchor ranging constraints as residual factors in a continuous-time factor graph. The maximum a posteriori estimate is then obtained by minimizing the sum of these whitened residuals within a sliding window using sparse nonlinear least squares. This design enables the framework to fully exploit asynchronous multi-sensor information, maintain a temporally continuous and smooth trajectory, and preserve high accuracy and robustness even when UWB anchors are scarce or ranging quality is degraded.

\section{Problem Formulation}

\begin{figure}[th]
	\centering \includegraphics[scale=0.22]{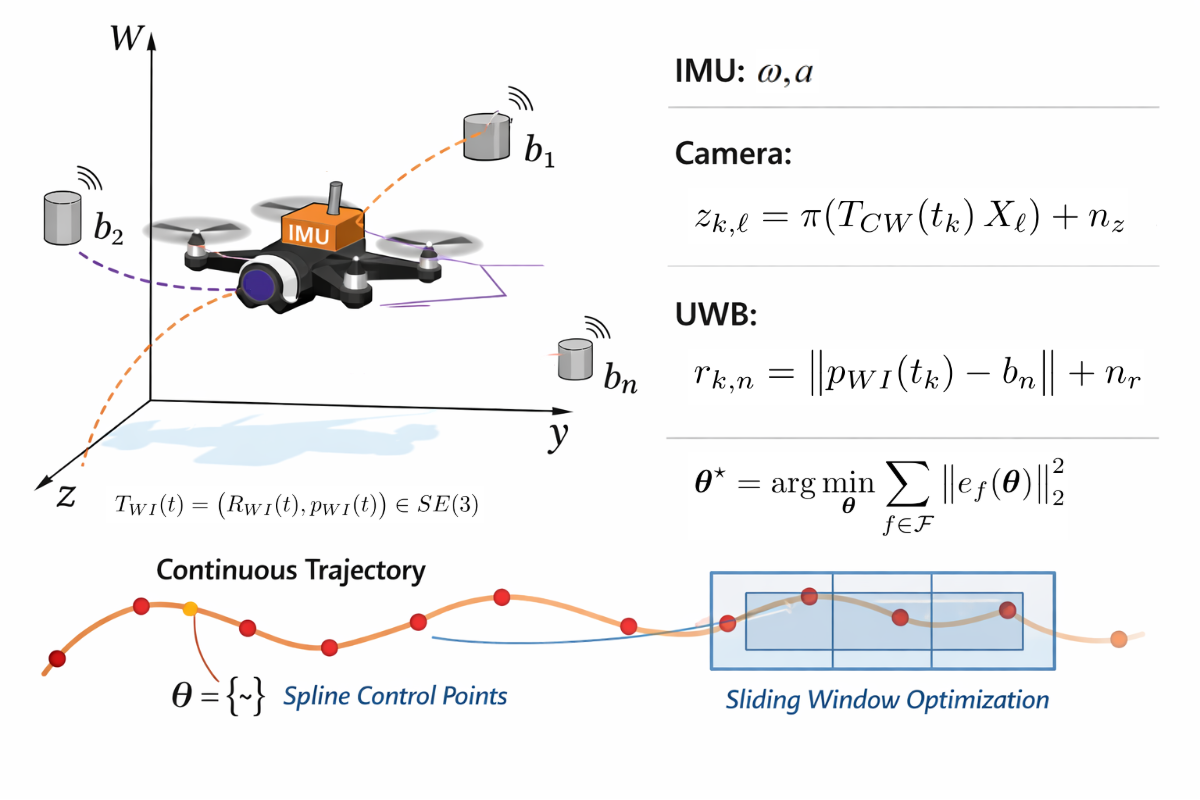}
	\caption{Overall sensing configuration of the VIR system.}
	\label{figure:2}
\end{figure}

We consider a mobile platform equipped with an IMU, a camera, and a UWB radio, as shown in Fig.~\ref{figure:2}.
Let $\mathcal{W}$, $\mathcal{I}$, and $\mathcal{C}$ denote the world, IMU, and camera frames, respectively.
A set of static ranging anchors (including physical anchors and later constructed virtual anchors) is deployed in $\mathcal{W}$,
with positions $\{b_n\}_{n=1}^{N}$, $b_n\in\mathbb{R}^3$.

We denote the continuous-time IMU pose with respect to $\mathcal{W}$ as
\begin{equation}
T_{WI}(t)=\big(R_{WI}(t),p_{WI}(t)\big)\in SE(3)
\end{equation} where $R_{WI}(t)\in SO(3)$ and $p_{WI}(t)\in\mathbb{R}^3$.
Let $T_{IC}$ be the known camera--IMU extrinsic calibration. Then the world-to-camera transform is
\begin{equation}
	T_{CW}(t)=T_{CI}T_{IW}(t),\quad T_{CI}=T_{IC}^{-1},\; T_{IW}(t)=T_{WI}(t)^{-1}
	\label{eq:tcw_def}
\end{equation}

At time $t_k$, the camera observes a landmark $X_\ell$ as
\begin{equation}
	z_{k,\ell}=\pi\!\left(T_{CW}(t_k)\,X_\ell\right)+n_z,
	\label{eq:vis_model}
\end{equation}
where $\pi(\cdot)$ denotes the camera projection model and $n_z$ is image noise.
For the $n$-th anchor at $b_n$, the UWB ranging measurement is modeled as
\begin{equation}
	r_{k,n}=\big\|p_{WI}(t_k)-b_n\big\|+n_r,
	\label{eq:rng_model}
\end{equation}
where $n_r$ captures measurement noise and possible NLOS-induced bias.
IMU measurements provide angular velocity and specific force, and are incorporated through standard preintegration factors.

Given time-stamped IMU measurements $\mathcal{U}$, visual observations $\mathcal{Z}$, and ranging measurements $\mathcal{R}$
over $[t_0,t_{\mathrm{end}}]$, we estimate the trajectory and associated variables by solving a MAP problem, which reduces to a
nonlinear least-squares form:
\begin{equation}
	\boldsymbol{\theta}^{\star}
	=\arg\min_{\boldsymbol{\theta}}
	\sum_{f\in\mathcal{F}}\big\|e_f(\boldsymbol{\theta})\big\|_2^{2},
	\label{eq:map_objective}
\end{equation}
where $\boldsymbol{\theta}$ collects the decision variables (e.g., spline control points, IMU biases, and landmark parameters),
and $e_f(\cdot)$ denotes a whitened residual associated with prior, IMU, visual, and ranging factors.

Direct optimization over the continuous trajectory $T_{WI}(t)$ is intractable. We therefore parameterize $T_{WI}(t)$ as a
finite-dimensional cubic B-spline in $SE(3)$ with uniformly spaced knots. All measurements are evaluated at their exact timestamps
as functions of the spline parameters, yielding a sparse nonlinear least-squares problem that is solved efficiently in a sliding-window
manner using standard sparse solvers.

\section{Methodology}

\subsection{Visual-Inertial Preprocessing}

Visual and inertial sensors provide complementary cues for short-term motion estimation.
We adopt a standard tightly-coupled VIO front-end that processes raw images and IMU data to produce pose priors
$\{R_{\mathrm{vio}}(t_k),\,p_{\mathrm{vio}}(t_k)\}$.
These priors are used to initialize the spline control points and to provide timestamp-aligned motion predictions for ranging consistency checks.
Once the back-end is running, the predictions are updated using the spline estimate from the previous sliding window, and the VIO priors can also be incorporated as soft prior factors when needed.

\subsection{Ranging Data Preprocessing}
\begin{figure}[th]
		\vspace{-0.5cm}
	\centering \includegraphics[scale=0.22]{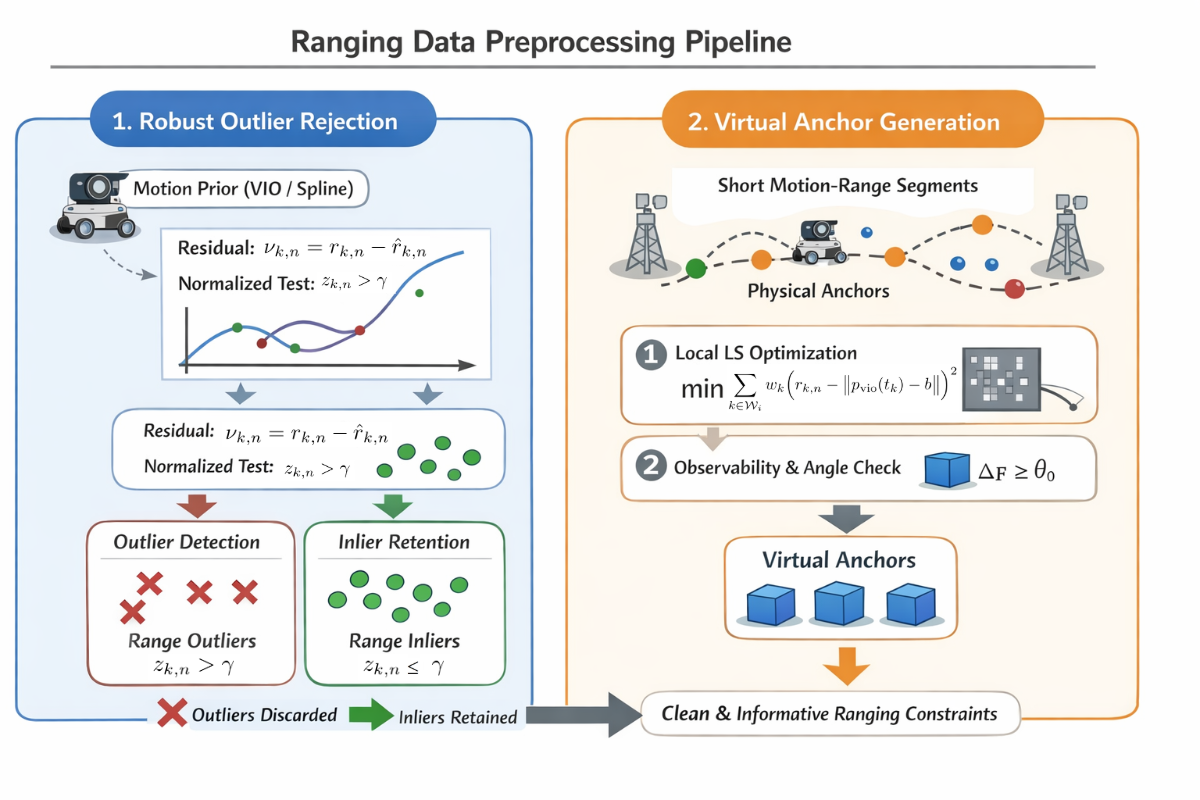}
	\protect\caption{Illustration of physical/virtual anchors and range inliers/outliers in 3D space.}
	\label{figure:3}
\end{figure}

UWB ranging is often degraded by NLOS propagation and multipath, resulting in biased, heavy-tailed errors.
With only a few anchors, multilateration on raw ranges is typically unreliable.
To provide clean and geometrically informative constraints for the subsequent continuous-time fusion (Figure~\ref{figure:3}),
we adopt a two-stage preprocessing pipeline: (i) robust outlier rejection, and (ii) virtual anchor construction
from short motion-range segments.
All positions and anchor coordinates in this subsection are expressed in the world frame $\mathcal{W}$.

\subsubsection*{1) Robust ranging outlier rejection}
Instead of fitting a local geometric model in the measurement space, we leverage the motion prior provided by the visual-inertial subsystem
(or the current spline estimate in the continuous-time back-end) to perform a statistically robust consistency check on ranging residuals.

For the $n$-th anchor (physical or virtual) at position $b_n$, let $p_{\mathrm{pr}}(t_k)$ be the predicted robot position at time $t_k$.
In our implementation, $p_{\mathrm{pr}}(t_k)$ is initialized by the VIO prior $p_{\mathrm{vio}}(t_k)$ and, once the back-end is running,
updated by the current spline estimate from the previous sliding window, i.e., $p_{\mathrm{pr}}(t_k)=p(t_k;\hat{\boldsymbol{\theta}})$,
where $p(t_k;\hat{\boldsymbol{\theta}})$ denotes the world-frame position evaluated from the continuous-time spline with
$\hat{\boldsymbol{\theta}}$.

The predicted range and innovation are
\begin{equation}
	\hat{r}_{k,n}=\big\|p_{\mathrm{pr}}(t_k)-b_n\big\|,\qquad
	\nu_{k,n}=r_{k,n}-\hat{r}_{k,n}
\end{equation}
where $r_{k,n}$ is the measured range. Over a sliding window
\begin{equation}
	\mathcal{W}_{i}=\{\,t_{k}\mid t_{k}\in[\bar{t}_{i}-\Delta,\;\bar{t}_{i}+\Delta]\,\}
\end{equation}
we compute the median and MAD of $\nu_{k,n}$:
\begin{equation}
	m_n^i=\operatorname{median}\{\nu_{k,n}\mid t_k\in\mathcal{W}_i\},\qquad	
\end{equation}
\begin{equation}
	s_n^i=\operatorname{median}\{|\nu_{k,n}-m_n^i|\mid t_k\in\mathcal{W}_i\}
\end{equation}
Each innovation is normalized as
\begin{equation}
	z_{k,n}=\frac{|\nu_{k,n}-m_n^i|}{s_n^i+\varepsilon}
\end{equation}
with $\varepsilon>0$ preventing division by zero. A measurement is treated as an outlier if
\begin{equation}
	z_{k,n}>\gamma
\end{equation}
where $\gamma$ is a tunable threshold. Outliers are discarded, and the remaining inlier ranges are used for virtual-anchor construction
and subsequent continuous-time estimation.

\subsubsection*{2) Virtual Anchor Generation}
When only a few physical anchors are available, multilateration remains unreliable even after outlier rejection.
To improve ranging geometry without additional infrastructure, we construct \emph{virtual anchors} (VAs) from short motion--range segments.
Different from propagating anchors via a single relative transform, we estimate each VA as a local anchor-like point by solving a small
windowed least-squares problem.

Consider the $n$-th physical anchor observed in the same short window $\mathcal{W}_i$.
Let $p_{\mathrm{vio}}(t_k)$ be the VIO position and $r_{k,n}$ the corresponding inlier range.
A candidate VA $b_n^i\in\mathbb{R}^3$ is obtained by
\begin{equation}
	b_{n}^{i}=\arg\min_{b\in\mathbb{R}^{3}}\sum_{k\in\mathcal{W}_{i}}w_{k}\Bigl(r_{k,n}-\bigl\|p_{\mathrm{vio}}(t_{k})-b\bigr\|\Bigr)^{2}
	\label{eq:va_ls}
\end{equation}
Here $w_k$ reflects range reliability; in our implementation, we use a residual-based reweighting to down-weight uncertain ranges, e.g.,
\begin{equation}
	w_k=\min\!\left(1,\frac{c}{|\nu_{k,n}|+\varepsilon}\right)
\end{equation}
where $\nu_{k,n}$ is the innovation defined in the outlier-rejection stage, and $c$ is a constant.

To retain informative and non-degenerate VAs, we evaluate their incremental observability contribution.
For a range measurement $\hat{r}_{k,b}=\|p_{\mathrm{vio}}(t_k)-b\|$ with variance $\sigma_r^2$, the Jacobian w.r.t. position is
\begin{equation}
	\mathbf{J}_{k,b}=\frac{\partial \hat{r}_{k,b}}{\partial p}
	=\frac{(p_{\mathrm{vio}}(t_k)-b)^{\top}}{\|p_{\mathrm{vio}}(t_k)-b\|}
	=\frac{(p_{\mathrm{vio}}(t_k)-b)^{\top}}{\hat{r}_{k,b}}
\end{equation}

We approximate the Fisher information of an anchor set $\mathcal{B}$ over $\mathcal{W}_i$ as
\begin{equation}
	F(\mathcal{B})=\sum_{k\in\mathcal{W}_i}\sum_{b\in\mathcal{B}}
	\mathbf{J}_{k,b}^{\top}\frac{1}{\sigma_r^{2}}\mathbf{J}_{k,b}
\end{equation}
Then we compute
\begin{equation}
	\Delta F=F(\mathcal{B}\cup\{b_{n}^{i}\})-F(\mathcal{B})
\end{equation}
and accept $b_n^i$ only if
\begin{equation}
	\lambda_{\min}(\Delta F)>\tau_{\lambda}
\end{equation}
We further enforce angular diversity by discarding candidates whose line-of-sight direction is nearly collinear with existing anchors.
Let $\bar{p}_i$ be the mean position in $\mathcal{W}_i$; we compute
\begin{equation}
	\theta_{\min}=\min_{b\in\mathcal{B}}\angle\bigl(b_{n}^{i}-\bar{p}_{i},\,b-\bar{p}_{i}\bigr)
\end{equation}
and reject $b_n^i$ if $\theta_{\min}<\theta_0$.

With robust outlier rejection followed by VA fitting and screening, we obtain a compact set of reliable ranging constraints.
In the subsequent continuous-time fusion, physical anchors and VAs are treated uniformly as static beacons providing range factors to
the spline trajectory. In particular, VAs are kept fixed after construction, and their uncertainty is absorbed by using a larger ranging
covariance for VA factors than that for physical-anchor factors.

\subsection{Continuous-time B-spline Trajectory Representation and Knot Placement}

\begin{figure}[th]
	\centering \includegraphics[scale=0.3]{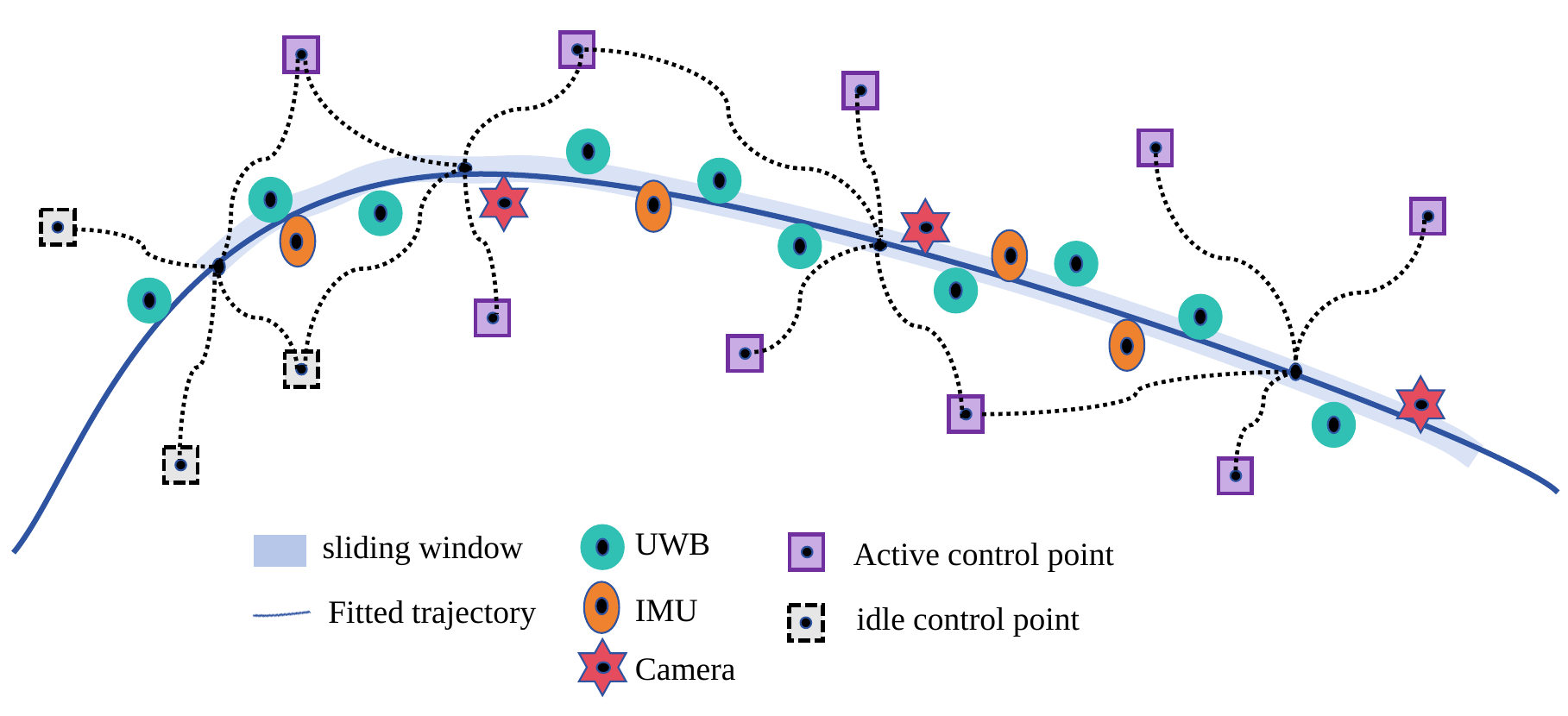}
	\protect\caption{Illustration of the continuous-time trajectory estimation.}
	\label{figure:4}
\end{figure}

In the continuous-time setting, the IMU pose in the world frame is represented by a trajectory
$T_{WI}(t)=(R_{WI}(t),p_{WI}(t))\in SE(3)$, where $p_{WI}(t)\in\mathbb{R}^3$ and $R_{WI}(t)\in SO(3)$.
For notational simplicity, we drop the subscripts in the following and denote
$T(t)\triangleq T_{WI}(t)$, $R(t)\triangleq R_{WI}(t)$, and $p(t)\triangleq p_{WI}(t)$.
We model $T(t)$ using a cubic B-spline parameterized by a finite set of translation and rotation control points.

\subsubsection*{1) Trajectory Representation}
Let $\{t_i\}_{i=0}^{M}$ be a uniform knot sequence for a cubic B-spline with spacing $h>0$,
\begin{equation}
	t_i=t_0+ih,\qquad i=0,\ldots,M .
\end{equation}
For $t\in[t_i,t_{i+1})$ with $i=0,\ldots,M-1$, define the normalized time
\begin{equation}
	u=\frac{t-t_i}{h}\in[0,1) .
\end{equation}
Let the cubic basis vector be
\begin{equation}
	\boldsymbol{\beta}(u)=\big[\beta_0(u),\,\beta_1(u),\,\beta_2(u),\,\beta_3(u)\big]^{\top} .
\end{equation}
The translation is represented by
\begin{equation}
	p(t)=\sum_{j=0}^{3}\beta_j(u)\,p_{i+j} ,
	\label{eq:bspline_pos}
\end{equation}
where $p_i\in\mathbb{R}^3$ are translation control points.

For rotation, we use a spline on $SO(3)$ via $\mathfrak{so}(3)$ control vectors $\phi_i\in\mathbb{R}^3$ and define, for $t\in[t_i,t_{i+1})$,
\begin{equation}
	R(t)=R_{i-1}\prod_{j=0}^{3}\exp\!\big(\beta_j(u)\,\phi_{i+j}^{\wedge}\big) ,
	\label{eq:bspline_rot}
\end{equation}
where $(\cdot)^{\wedge}$ maps $\mathbb{R}^3$ to $\mathfrak{so}(3)$ and $(\cdot)^{\vee}$ denotes its inverse.
With this representation, $p(t)$ and $R(t)$ (and their time derivatives) can be evaluated analytically.
In particular, the body (IMU-frame) angular velocity is
\begin{equation}
	\omega(t)=\big(R^{\top}(t)\dot{R}(t)\big)^{\vee} .
\end{equation}

\subsubsection*{2) Uniform Knot Placement}
We adopt a uniform knot placement strategy, i.e., a constant knot interval $h$ over the entire trajectory.
Given a time span $[t_{0}, t_{\mathrm{end}}]$, the number of knots is
\begin{equation}
	M+1 = 1 + \bigg\lceil \frac{t_{\mathrm{end}}-t_{0}}{h} \bigg\rceil ,
\end{equation}
so that $t_M=t_0+Mh \ge t_{\mathrm{end}}$.
The interval $h$ trades approximation fidelity for computational cost: smaller $h$ introduces more control points
and better captures fast motion, while larger $h$ reduces variables at the expense of increased approximation error.
In practice, $h$ is chosen according to the expected motion dynamics and sensor sampling rates (e.g., ensuring
multiple IMU samples per knot interval and sufficient visual/ranging constraints across adjacent intervals),
yielding a simple and analytically tractable continuous-time representation for the subsequent factor-graph estimation.

\subsection{Continuous-time VIR Factor-Graph Formulation}
\label{subsec:factor_graph}

Based on the continuous-time B-spline trajectory representation, we formulate state estimation
as a nonlinear least-squares problem over the spline control points and a set of auxiliary parameters
(e.g., IMU biases and landmark positions). All measurements are encoded as factors in a factor graph.

Let $\boldsymbol{\theta}$ collect all spline control points for translation and rotation, as well as IMU bias
parameters $\mathbf{b}_{a},\mathbf{b}_{g}$ and (if estimated) 3D landmark positions. In our experiments,
anchors are treated as fixed beacons, including the constructed virtual anchors.
Given IMU measurements $\mathcal{U}$, visual observations $\mathcal{Z}$, and ranging measurements $\mathcal{R}$,
the maximum a posteriori estimate is obtained by
\begin{equation}
	\begin{aligned}
		\boldsymbol{\theta}^{\star}
		&=\arg\min_{\boldsymbol{\theta}} \; J(\boldsymbol{\theta}), \\
		J(\boldsymbol{\theta})
		&=\sum_{f\in\mathcal{F}_{\text{prior}}}\!\!\|e_f(\boldsymbol{\theta})\|_2^{2}
		+\sum_{f\in\mathcal{F}_{\text{imu}}}\!\!\|e_f(\boldsymbol{\theta})\|_2^{2} \\
		&\quad+\sum_{f\in\mathcal{F}_{\text{vis}}}\!\!\|e_f(\boldsymbol{\theta})\|_2^{2}
		+\sum_{f\in\mathcal{F}_{\text{rng}}}\!\!\|e_f(\boldsymbol{\theta})\|_2^{2}
	\end{aligned}
	\label{eq:overall_cost}
\end{equation}

where each $e_f(\cdot)$ denotes a whitened residual associated with a measurement factor, with whitening matrices
defined by the corresponding measurement covariances. Below we briefly detail the main factor types.

\subsubsection*{1) IMU factor}
Let $(\omega_{k}^{m},a_{k}^{m})$ denote the IMU measurements at time $t_{k}$, where
$\omega_{k}^{m}\in\mathbb{R}^3$ and $a_{k}^{m}\in\mathbb{R}^3$ are the measured angular velocity
and specific force expressed in the IMU frame.
Given the continuous-time pose $T(t)=(R(t),p(t))$ and its derivatives from the B-spline, the predicted body angular velocity
and specific force are
\begin{equation}
	\omega_{k}(\boldsymbol{\theta})=\omega(t_k;\boldsymbol{\theta}),\qquad
	a_{k}(\boldsymbol{\theta})=R^{\top}(t_{k})\big(\ddot{p}(t_{k})-g^{W}\big)
\end{equation}
where $g^{W}$ is the gravity vector in the world frame, and the body angular velocity is obtained from
the spline rotation as $\omega(t)=\big(R^{\top}(t)\dot{R}(t)\big)^{\vee}$.
The IMU measurement model is
\begin{align}
	\omega_{k}^{m} & =\omega_{k}(\boldsymbol{\theta})+\mathbf{b}_{g}+n_{g}\\
	a_{k}^{m} & =a_{k}(\boldsymbol{\theta})+\mathbf{b}_{a}+n_{a}
\end{align}
with zero-mean Gaussian noises $n_{g}$ and $n_{a}$. The corresponding whitened residual is
\begin{equation}
	e_{k}^{\text{imu}}(\boldsymbol{\theta})=
	\Sigma_{\text{imu}}^{-\frac{1}{2}}
	\begin{bmatrix}
		\omega_{k}^{m}-\omega_{k}(\boldsymbol{\theta})-\mathbf{b}_{g}\\
		a_{k}^{m}-a_{k}(\boldsymbol{\theta})-\mathbf{b}_{a}
	\end{bmatrix}
	\label{eq:imu_factor}
\end{equation}
where $\Sigma_{\text{imu}}=\mathrm{diag}(\Sigma_{\omega},\Sigma_{a})\in\mathbb{R}^{6\times6}$
denotes the IMU noise covariance used for whitening.

\subsubsection*{2) Visual reprojection factor}
Let $X_{\ell}\in\mathbb{R}^{3}$ be the 3D position of landmark $\ell$ in the world frame,
and $z_{k,\ell}$ the image observation at time $t_{k}$.
Let $T_{WI}(t_k)$ denote the IMU pose evaluated from the spline, and let $T_{IC}$ be the fixed extrinsic
transform from the camera frame to the IMU frame. Then $T_{CI}=T_{IC}^{-1}$, and the world-to-camera transform is
\begin{equation}
	T_{CW}(t_k)=T_{CI}\,T_{IW}(t_k),
\end{equation}
where $T_{IW}(t_k)=T_{WI}^{-1}(t_k)$.
The predicted pixel coordinate is
\begin{equation}
	\hat{z}_{k,\ell}(\boldsymbol{\theta})
	=\pi\!\left(T_{CW}(t_{k};\boldsymbol{\theta})\,X_{\ell}\right)
\end{equation}
where $\pi(\cdot)$ is the camera projection model (including intrinsics).
The visual residual is given by
\begin{equation}
	e_{k,\ell}^{\text{vis}}(\boldsymbol{\theta})
	=\Sigma_{\text{vis}}^{-\frac{1}{2}}\big(z_{k,\ell}-\hat{z}_{k,\ell}(\boldsymbol{\theta})\big)
	\label{eq:vis_factor}
\end{equation}
where $\Sigma_{\text{vis}}$ denotes the image measurement covariance (in pixel units) used for whitening.

\subsubsection*{3) Ranging factor for physical and virtual anchors}
For the $n$-th anchor (physical or virtual) with position $b_{n}$ in the world frame,
let $r_{k,n}$ be the (inlier) ranging measurement at time $t_{k}$.
The predicted range is
\begin{equation}
	\hat{r}_{k,n}(\boldsymbol{\theta})=\big\| p(t_{k};\boldsymbol{\theta})-b_{n}\big\|
\end{equation}
and the measurement model is
\begin{equation}
	r_{k,n}=\hat{r}_{k,n}(\boldsymbol{\theta})+n_{r}
\end{equation}
where $n_{r}\sim\mathcal{N}(0,\sigma_r^2)$ is a zero-mean Gaussian noise.
The corresponding whitened residual reads
\begin{equation}
	e_{k,n}^{\text{rng}}(\boldsymbol{\theta})
	=\sigma_r^{-1}\big(r_{k,n}-\hat{r}_{k,n}(\boldsymbol{\theta})\big)
	\label{eq:rng_factor}
\end{equation}
which is equivalent to using $\Sigma_{\text{rng}}=\sigma_r^2$ for whitening.
Physical and virtual anchors enter the formulation in exactly the same way; the latter are additional static points
$b_n$ generated by the preprocessing described in Section~4.2.
In practice, we use a larger ranging variance for virtual-anchor factors to account for their fitting uncertainty and local-model mismatch.

\smallskip
With these factor definitions, the overall problem \eqref{eq:overall_cost} becomes a standard nonlinear least-squares optimization
over the spline control points and auxiliary variables, which can be efficiently solved in a sliding-window fashion using sparse linear solvers,
with marginalization of old variables yielding a prior factor on the remaining states.

\section{Experimental Evaluation}

This section presents the experimental evaluation of the proposed
method. We validate our approach on both public benchmark datasets
and real-world flight experiments. All experiments are run on a workstation
equipped with an Intel(R) Core(TM) i9-10900K CPU @ 3.70 GHz, 64 GB
RAM, and Ubuntu 20.04. We use the root mean square error (RMSE) of
the absolute trajectory error (ATE) as the primary evaluation metric.
For each dataset, we conduct 10 trials using the default calibrations
and ATE RMSE per run, report the median over 10 runs.

\subsection{Implementation and Parameter Settings}

Unless otherwise stated, the knot interval $h$, sliding-window length, outlier-rejection parameters $(\Delta,\gamma,\varepsilon)$,
and VA screening thresholds $(\tau_{\lambda},\theta_{0})$ are fixed across all experiments; the settings are summarized in Table~1.

\begin{table}[!t]
	\caption{Default parameter settings used in experiments.}
	\label{Tab1}
	\centering
	\renewcommand{\arraystretch}{1.15}
	\setlength{\tabcolsep}{5pt}
	\begin{tabular}{p{2.5cm} p{2.5cm} p{2.6cm}}
		\toprule
		\textbf{Parameter} & \textbf{Default} & \textbf{Range} \\
		\midrule
		Knot interval $h$ 
		& $0.05\,\mathrm{s}$ 
		& $[0.03,\,0.10]\,\mathrm{s}$ \\
		
		Window length 
		& $4.0\,\mathrm{s}$ 
		& $[3,\,6]\,\mathrm{s}$ \\
		
		\multirow[t]{3}{*}{Outlier rejection} 
		& $\Delta=0.5\,\mathrm{s}$ 
		& $\Delta\in[0.3,\,1.0]\,\mathrm{s}$ \\
		& $\gamma=3.5$ 
		& $\gamma\in[3.0,\,4.5]$ \\
		& $\varepsilon=10^{-3}\,\mathrm{m}$ 
		& $\varepsilon\in[10^{-4},\,10^{-2}]\,\mathrm{m}$ \\
		
		\multirow[t]{2}{*}{VA screening}
		& $\tau_\lambda=0.05$ 
		& $\tau_\lambda\in[0.01,\,0.20]$ \\
		& $\theta_0=15^\circ$ 
		& $\theta_0\in[10^\circ,\,25^\circ]$ \\
		
		\multirow[t]{2}{*}{Ranging noise}
		& $\sigma_r=0.10\,\mathrm{m}$ 
		& $\sigma_r\in[0.05,\,0.20]\,\mathrm{m}$ \\
		& $\sigma_{va}=0.20\,\mathrm{m}$ 
		& $\sigma_{va}\in[0.15,\,0.40]\,\mathrm{m}$ \\
		
		Solver / iterations 
		& LM / $5$--$10$ iters 
		& per window \\
		\bottomrule
	\end{tabular}
\end{table}

\subsection{Experimental Evaluation in Public Datasets}

\subsubsection*{1) Synthetic EuRoC MAV Dataset}

In this experiment, we evaluate the proposed method on the EuRoC MAV
dataset. Since EuRoC does not provide UWB measurements, we simulate
UWB ranging data based on the ground-truth trajectories. The UWB anchors
are defined in the ground-truth world frame. For each room (MH, V1,
V2), we place four static anchors near the corners of the trajectory
bounding box at a fixed height and generate synthetic range measurements
by adding Gaussian noise and occasional positive outliers to the ground-truth
distances (corresponding to position perturbations of approximately
10, 10, and 50 cm in the x, y, and z directions, respectively).
The three-anchor configuration (A=3) is obtained by removing one anchor
from the four-anchor setup, so that both configurations share the
same underlying geometry.

We compare the proposed method with several recent works, covering
both continuous-time and discrete-time fusion frameworks. Since there
is currently no continuous-time fusion method specifically designed
for VIO-UWB, we select two closely related continuous-time approaches
as baselines: a spline-based IMU-UWB system (Spline-UI) and a spline-based
VIO method (Spline-VIO). In the discrete-time setting, we consider
three VIO-UWB fusion methods as baselines: EKF-VIU, HCCNet, and
Refloc. In addition, we explicitly study two anchor configurations:
full anchor deployment (A=4) and a missing-anchor scenario (A=3).

\begin{figure}
	\setlength{\abovecaptionskip}{-0.0cm}
	\centering
	\includegraphics[scale=0.18]{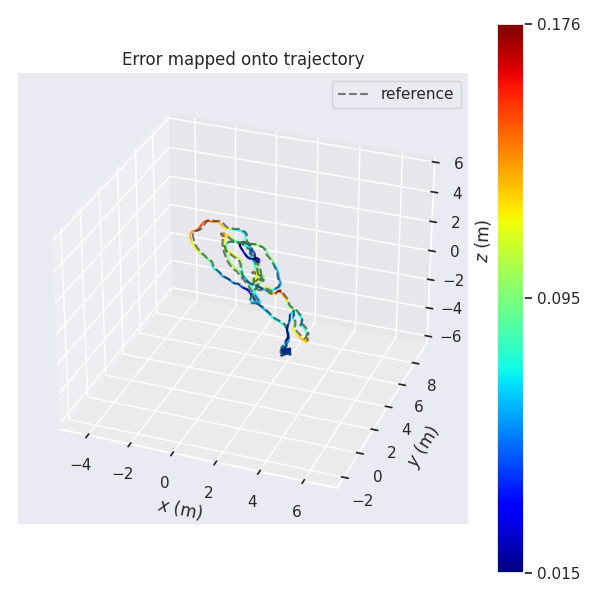} 
\includegraphics[scale=0.18]{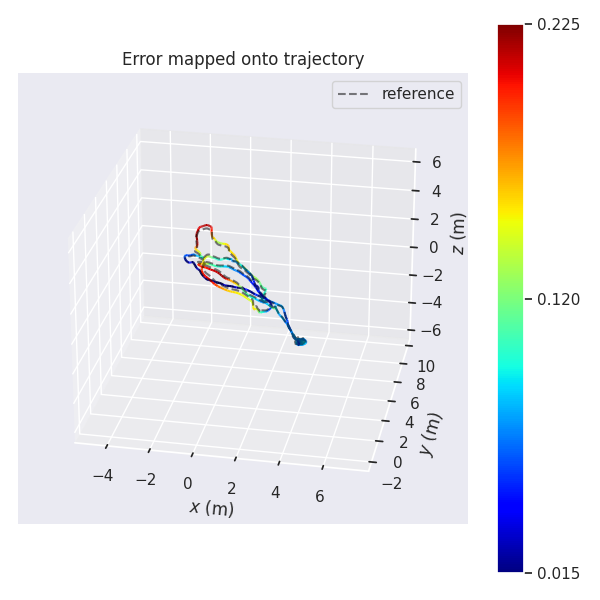}
\includegraphics[scale=0.18]{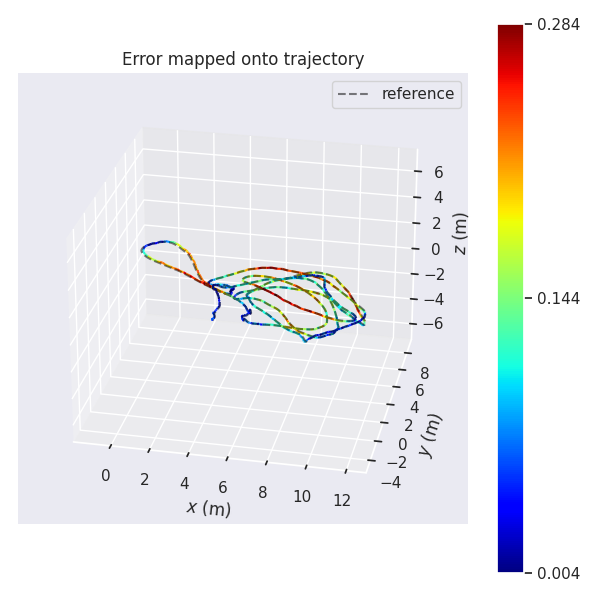}
\includegraphics[scale=0.18]{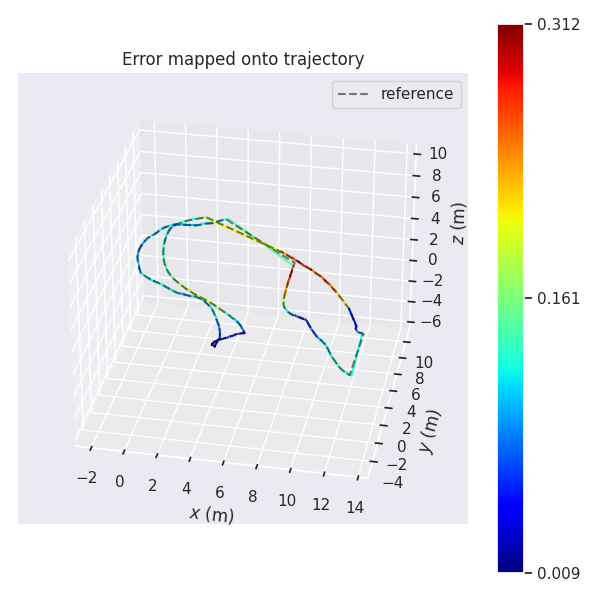} 
\includegraphics[scale=0.18]{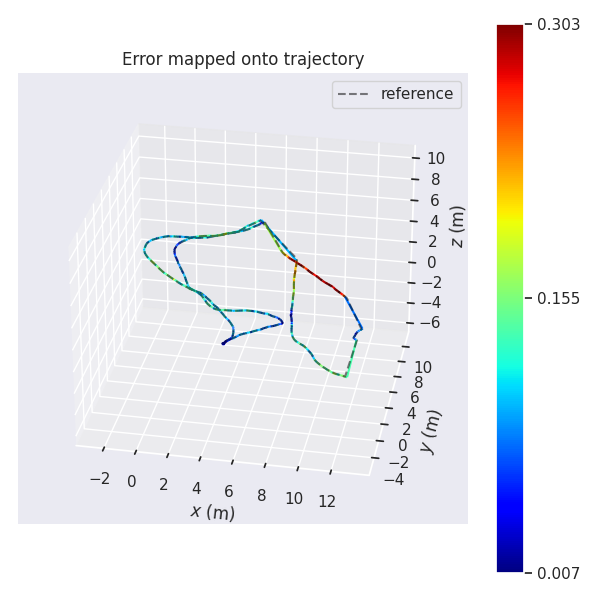}
\includegraphics[scale=0.18]{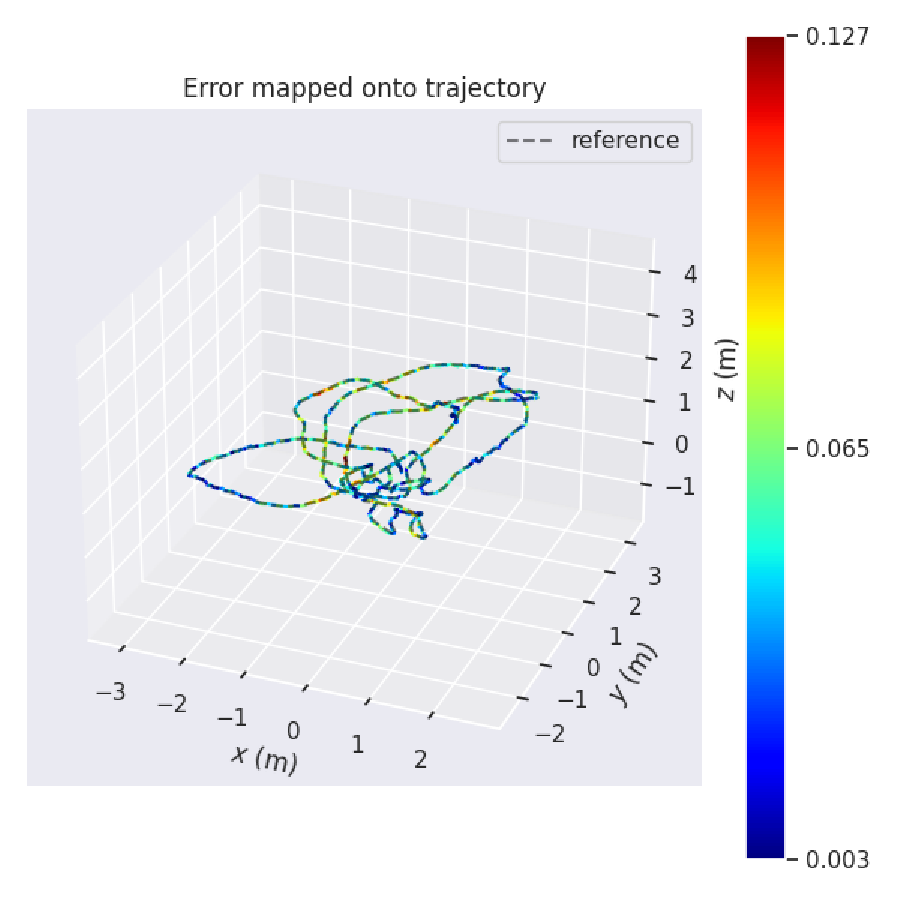}
\includegraphics[scale=0.18]{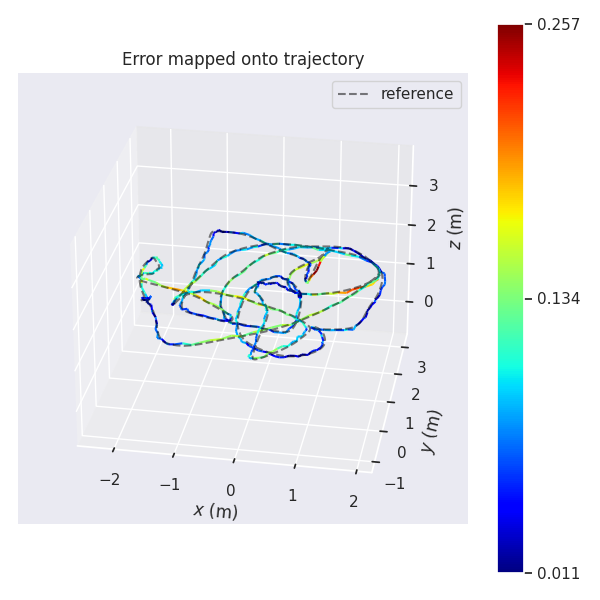}
\includegraphics[scale=0.18]{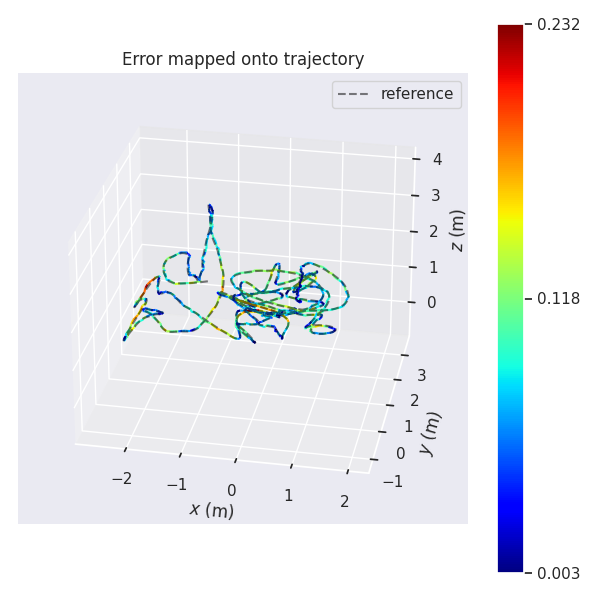}
\includegraphics[scale=0.18]{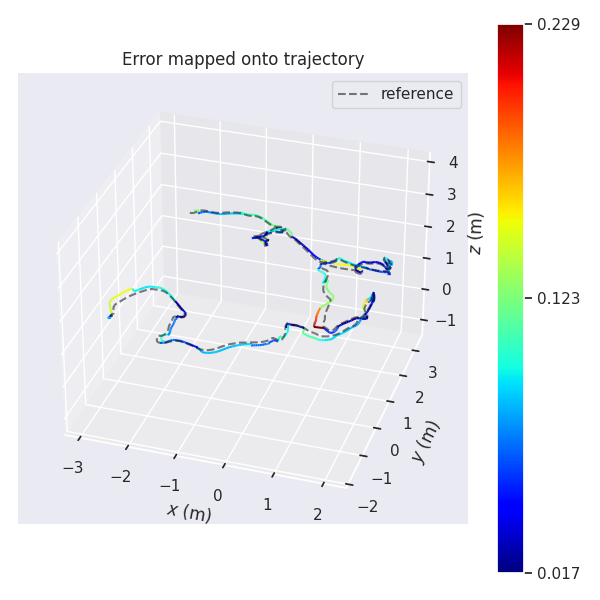}
\includegraphics[scale=0.18]{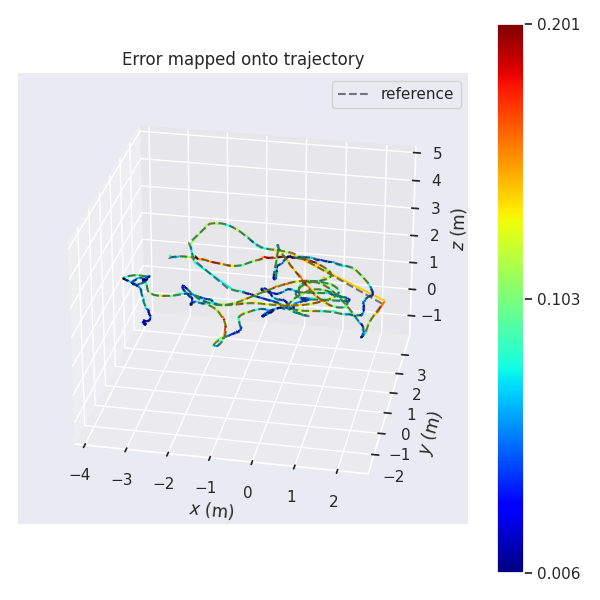}
\includegraphics[scale=0.18]{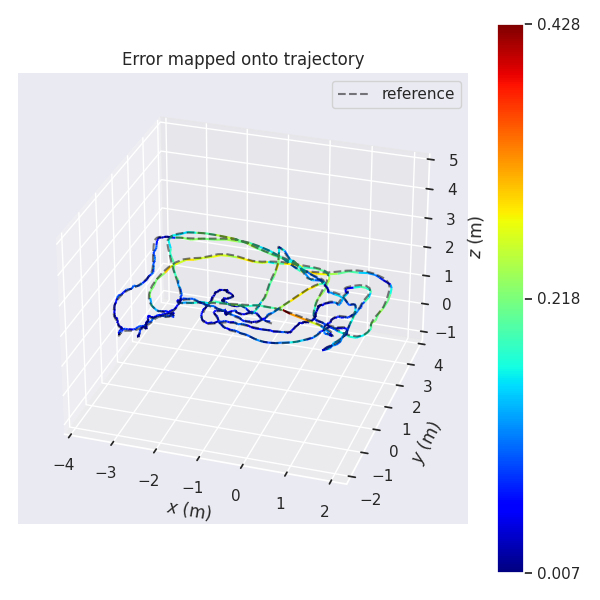}	
	
	\vspace{0.1cm}
	\caption{Trajectory estimation on the Euroc sequences.}
	\label{figure:6}
\end{figure}

\begin{table}[!h]
	\caption{Comparison of localization accuracy on EuRoC dataset.}
	\label{Tab2}
	\centering
	\footnotesize
	\setlength{\tabcolsep}{3.2pt}
	\begin{tabular}{lccccccc}
		\toprule
		\multirow{2}{*}{Sequence} & \multicolumn{2}{c}{Continuous-time} & \multicolumn{3}{c}{Discrete-time} & \multicolumn{2}{c}{Proposed} \\
		\cmidrule(lr){2-3}\cmidrule(lr){4-6}\cmidrule(lr){7-8}
		& CT-UI & CT-VIO & EKF-VIU & HCCNet & Refloc & Our* & Our \\
		\midrule
		MH\_01 & 0.205 & 0.070 & 0.150 & 0.061 & 0.116 & 0.068 & \textbf{0.061}\\
		MH\_02 & 0.273 & 0.061 & 0.094 & 0.065 & 0.062 & 0.073 & \textbf{0.060}\\
		MH\_03 & 0.288 & 0.153 & 0.146 & 0.128 & 0.129 & 0.137 & \textbf{0.113}\\
		MH\_04 & 0.269 & 0.149 & 0.136 & 0.119 & 0.103 & 0.126 & \textbf{0.084}\\
		MH\_05 & 0.238 & 0.148 & 0.118 & 0.116 & 0.100 & 0.124 & \textbf{0.094}\\
		\midrule
		V1\_01 & 0.193 & 0.106 & 0.082 & \textbf{0.059} & 0.062 & 0.091 & 0.060\\
		V1\_02 & 0.304 & failed & 0.116 & 0.097 & 0.094 & 0.144 & \textbf{0.096}\\
		V1\_03 & 0.296 & failed & 0.117 & \textbf{0.097} & 0.125 & 0.171 & 0.103\\
		\midrule
		V2\_01 & 0.137 & 0.110 & 0.101 & \textbf{0.044} & 0.069 & 0.092 & 0.051\\
		V2\_02 & 0.235 & 0.126 & 2.067 & 0.086 & 0.102 & 0.114 & \textbf{0.082}\\
		V2\_03 & 0.213 & failed & 0.117 & 0.121 & 0.103 & 0.154 & \textbf{0.094}\\
		\midrule
		Average & 0.241 & 0.115 & 0.295 & 0.090 & 0.094 & 0.118 & \textbf{0.082}\\
		\bottomrule
	\end{tabular}
\end{table}

Table~2 reports the absolute trajectory RMSE (m) on
the EuRoC MH, V1, and V2 sequences for all compared methods. As visualized in Figures 5 and 6, we also provide qualitative trajectory comparisons on the MH and V1/V2 sequences. Overall,
the proposed continuous-time VIR fusion with
four anchors (A=4) achieves the best average accuracy of 0.0815 m,
significantly improving over the continuous-time baselines Spline-UI and Spline-VIO, and also outperforming the strongest
discrete-time baseline HCCNet. The three-anchor variant
(A=3) is slightly less accurate than A=4, as expected from the reduced
ranging geometry, but remains competitive and generally superior to
Spline-UI and EKF-VIU on most sequences. In addition, several baselines
either diverge or fail on challenging sequences (e.g., Spline-VIO
on V1\_02, V1\_03, V2\_03
and EKF-VIU on V2\_02), whereas the proposed method
produces stable estimates across all sequences, demonstrating both
higher accuracy and better robustness under few-anchor configurations.

\subsubsection*{2) Synthetic UZH-FPV dataset}

In this experiment, we further evaluate the proposed method on the
UZH-FPV dataset. Similar to the EuRoC setting, we synthetically generate
UWB range measurements from the ground-truth trajectories and add
them to the dataset. Unless otherwise stated, the synthetic generation follows the same update rate, noise level, and outlier setting as in EuRoC.
The proposed approach is then compared against
several recent continuous-time and discrete-time fusion baselines.

\begin{figure}[h]
	\setlength{\abovecaptionskip}{-0.0cm} \centering \includegraphics[scale=0.18]{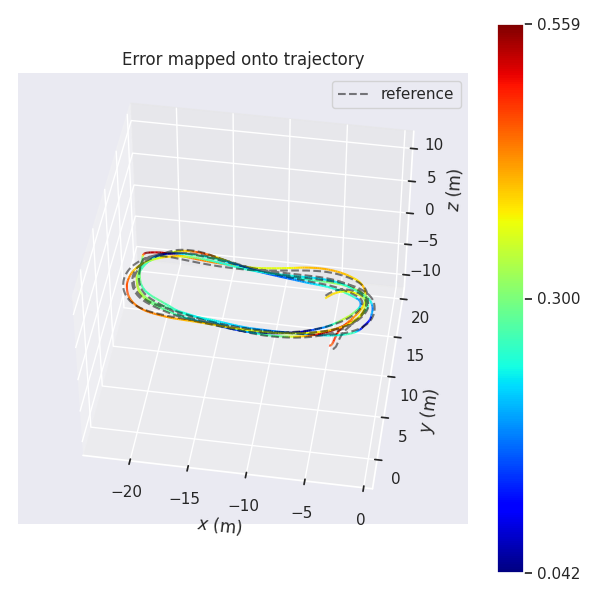}
	\includegraphics[scale=0.18]{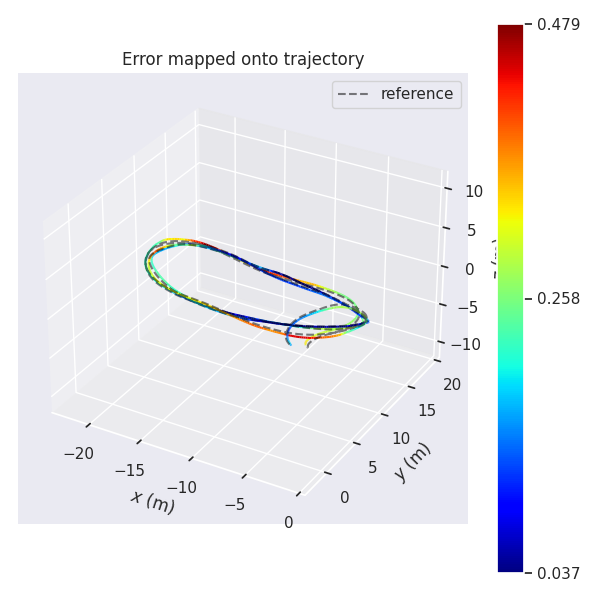} 
	\includegraphics[scale=0.18]{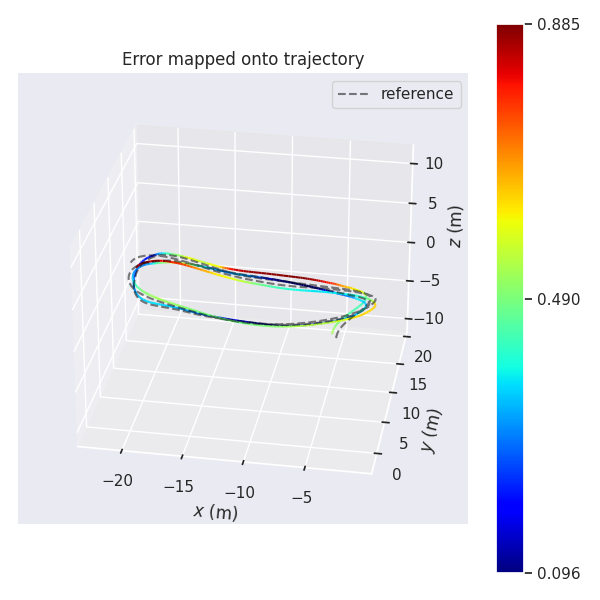}
	\includegraphics[scale=0.18]{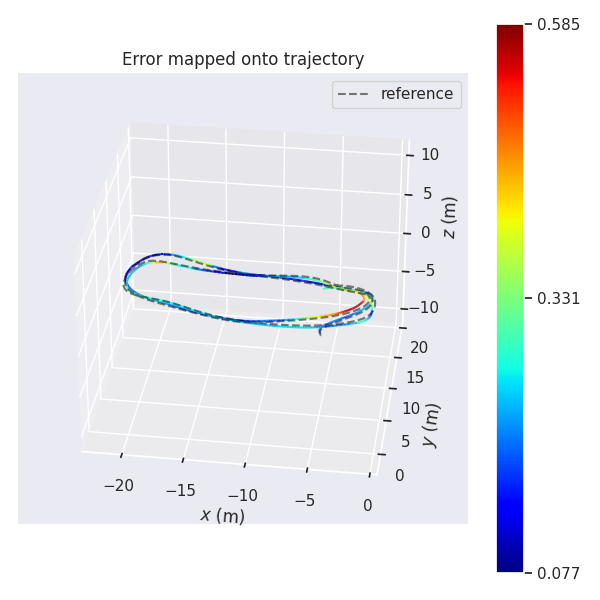}
	\includegraphics[scale=0.18]{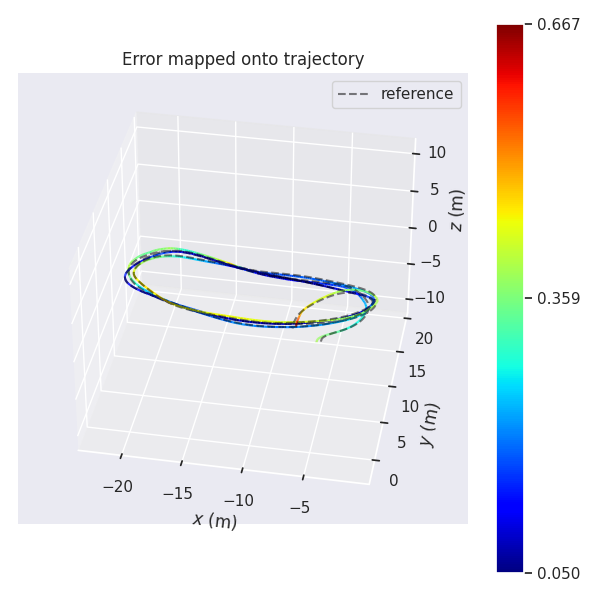} 
	\includegraphics[scale=0.18]{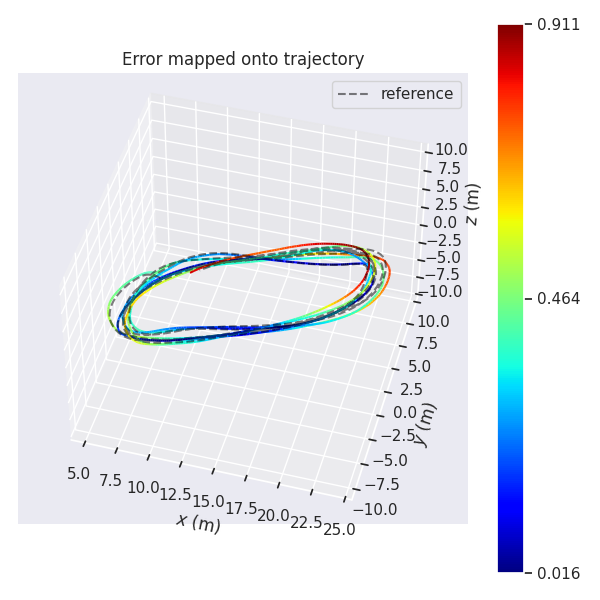}
	\includegraphics[scale=0.18]{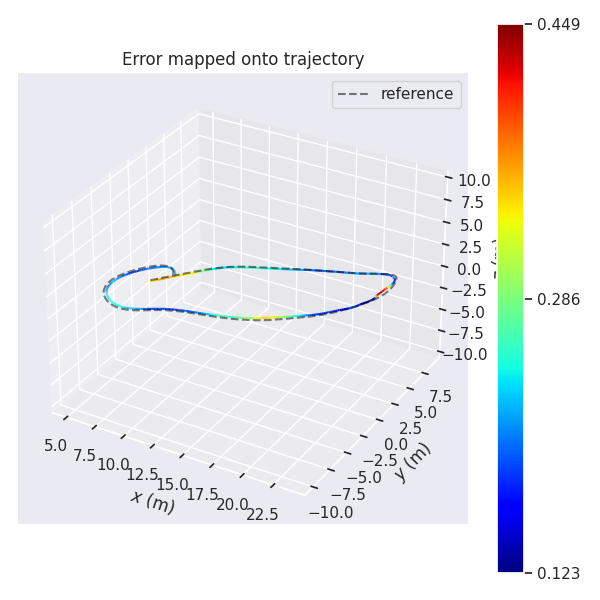}
	\includegraphics[scale=0.18]{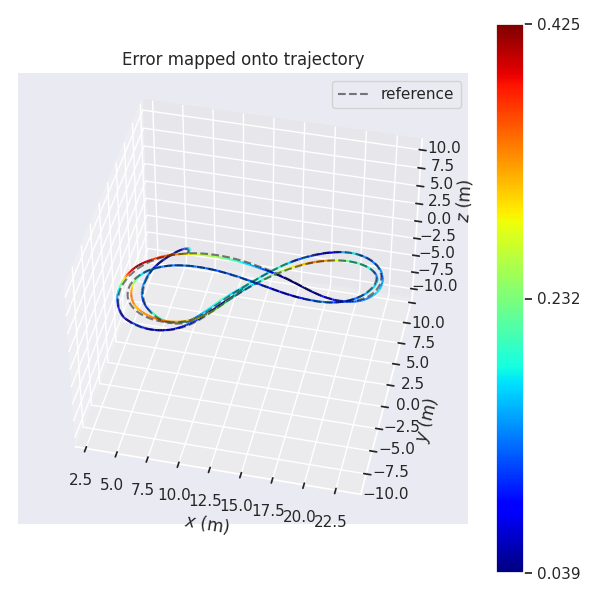}
    \includegraphics[scale=0.18]{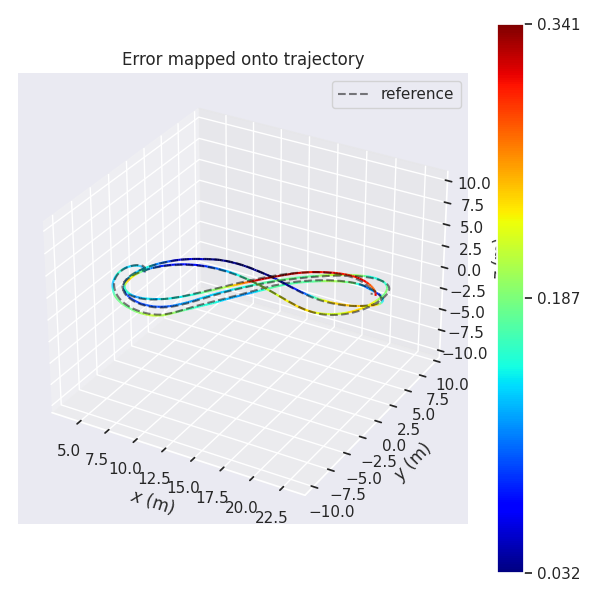}	
	
	 \vspace{0.1cm}
	
	\caption{Trajectory estimation on the UZH-FPV indoor 45/forward sequences.}

	\label{figure:7}
\end{figure}

\begin{table}[!h]
	\caption{Comparison of localization accuracy on UZH-FPV dataset.}
	\label{Tab3}
	\centering
	\footnotesize
	\setlength{\tabcolsep}{3.2pt}
	\begin{tabular}{lccccccc}
		\toprule
		\multirow{2}{*}{Sequence} & \multicolumn{2}{c}{Continuous-time} & \multicolumn{3}{c}{Discrete-time} & \multicolumn{2}{c}{Proposed}\\
		\cmidrule(lr){2-3}\cmidrule(lr){4-6}\cmidrule(lr){7-8}
		& CT-UI & CT-VIO & EKF-VIU & HCCNet & Refloc & Our* & Our \\
		\midrule
		in\_45\_2  & 1.876 & 1.364 & 0.636 & 0.424 & 0.471 & 0.556 & \textbf{0.346}\\
		in\_45\_4  & 1.931 & 1.796 & 0.624 & 0.405 & 0.394 & 0.514 & \textbf{0.304}\\
		in\_45\_9  & 1.610 & 1.422 & 0.593 & 0.501 & 0.555 & 0.622 & \textbf{0.476}\\
		in\_45\_12 & failed & 1.279 & 0.493 & 0.401 & 0.388 & 0.533 & \textbf{0.306}\\
		in\_45\_13 & 1.968 & failed & 0.510 & 0.416 & 0.404 & 0.544 & \textbf{0.315}\\
		\midrule
		in\_for\_3  & 1.343 & 0.994 & 0.452 & \textbf{0.416} & 0.443 & 0.581 & 0.422\\
		in\_for\_5  & failed & 0.876 & 0.297 & 0.287 & 0.303 & 0.484 & \textbf{0.246}\\
		in\_for\_9  & 1.344 & 0.961 & 0.336 & 0.251 & 0.234 & 0.340 & \textbf{0.193}\\
		in\_for\_10 & 1.289 & 0.885 & 0.283 & 0.257 & 0.237 & 0.307 & \textbf{0.164}\\
		\midrule
		Average & 1.623 & 1.197 & 0.469 & 0.373 & 0.392 & 0.498 & \textbf{0.308}\\
		\bottomrule
	\end{tabular}
\end{table}

Table~3 summarizes the localization performance on
the UZH-FPV dataset. The proposed continuous-time VIR
fusion method with four anchors (A=4) achieves the lowest average ATE of
0.3081 m, clearly outperforming the continuous-time baselines Spline-UI
and Spline-VIO, as well as the discrete-time
fusion methods EKF-VIU, HCCNet, and Refloc. As shown in Figure~\ref{figure:7}, we further provide trajectory visualizations for qualitative comparison. On most sequences, the four-anchor configuration either attains
the best result or remains very close to the best baseline. The three-anchor
variant (A=3) exhibits slightly higher errors,
which is consistent with the reduced ranging geometry, but still improves
considerably over Spline-UI and Spline-VIO. In addition, Spline-UI and Spline-VIO
fail on several challenging sequences (e.g., indoor\_45\_12,
indoor\_45\_13, indoor\_forward\_5),
whereas the proposed method produces stable estimates across all runs.
These observations suggest that the proposed continuous-time fusion
framework can provide more accurate and robust localization, particularly
under realistic few-anchor indoor conditions.

\subsubsection*{3) NTU VIRAL dataset}

In this section, we evaluate the performance of the proposed fusion
framework on the NTU VIRAL dataset, which provides synchronized
camera, inertial measurement unit (IMU), and ultra-wideband (UWB)
data. The UWB setup is configured as follows: the UAV is equipped
with two Humatics P440 UWB radio modules (IDs 200 and 201), each connected
to two antennas (A and B), resulting in four on-board ranging nodes
(200.A, 200.B, 201.A, 201.B). Three UWB anchors (IDs 100, 101, 102)
are deployed on the ground.

To investigate the impact of anchor availability, we construct two
anchor configurations: a full-anchor setup using all three ground
anchors (A=3), and a degraded few-anchor setup using only two anchors
(A=2). The latter is obtained by filtering out all range measurements
associated with one of the anchors from the original logs, while keeping
all other sensor data and algorithmic settings unchanged. This allows
us to emulate anchor failures or sparse UWB deployments, and to systematically
study the robustness of the proposed continuous-time fusion framework
under reduced geometric constraints.

\begin{table}[!h]
	\caption{Comparison of localization accuracy on NTU VIRAL dataset.}
	\label{Tab4}
	\centering
	\footnotesize
	\setlength{\tabcolsep}{3.2pt}
	\begin{tabular}{lccccccc}
		\toprule
		\multirow{2}{*}{Sequence} & \multicolumn{2}{c}{Continuous-time} & \multicolumn{3}{c}{Discrete-time} & \multicolumn{2}{c}{Proposed} \\
		\cmidrule(lr){2-3}\cmidrule(lr){4-6}\cmidrule(lr){7-8}
		& CT-UI & CT-VIO & EKF-VIU & HCCNet & Refloc & Our* & Our \\
		\midrule
		eee\_01 & 0.484 & 0.575 & 0.232 & 0.201 & 0.227 & 0.257 & \textbf{0.166}\\
		eee\_02 & 0.407 & 0.483 & 0.204 & 0.187 & 0.174 & 0.225 & \textbf{0.143}\\
		eee\_03 & 0.397 & 0.463 & 0.186 & 0.171 & 0.189 & 0.249 & \textbf{0.118}\\
		\midrule
		nya\_01 & 0.366 & 0.356 & 0.208 & 0.186 & 0.193 & 0.273 & \textbf{0.157}\\
		nya\_02 & 0.389 & 0.402 & 0.193 & \textbf{0.160} & 0.185 & 0.224 & 0.179\\
		nya\_03 & 0.731 & 0.769 & 0.206 & 0.193 & \textbf{0.175} & 0.309 & 0.184\\
		\midrule
		sbs\_01 & 0.452 & 0.488 & 0.253 & 0.207 & 0.197 & 0.251 & \textbf{0.152}\\
		sbs\_02 & 0.488 & 0.513 & 0.219 & 0.184 & 0.173 & 0.225 & \textbf{0.134}\\
		sbs\_03 & 0.421 & 0.631 & 0.185 & 0.180 & 0.183 & 0.263 & \textbf{0.160}\\
		\midrule
		Average & 0.459 & 0.520 & 0.210 & 0.159 & 0.189 & 0.253 & \textbf{0.155}\\
		\bottomrule
	\end{tabular}
\end{table}

Table 4 reports the ATE RMSE on the NTU VIRAL sequences
for all compared methods. Overall, the proposed continuous-time VIR
fusion with three anchors (A=3) achieves the best average accuracy
of 0.1548 m, slightly outperforming the strongest discrete-time baseline
HCCNet and clearly improving over EKF-VIU and
Refloc, as well as the continuous-time baselines Spline-UI and Spline-VIO. On seven out of nine sequences
(eee\_01--03, nya\_01, sbs\_01--03), A=3 yields the lowest error,
while on the remaining two sequences (nya\_02 and nya\_03) it remains
close to the best learning-based methods. The two-anchor configuration
(A=2) shows a moderate degradation to 0.2530 m on average, which is
consistent with the reduced geometric constraints, but still provides
substantial gains over Spline-UI and Spline-VIO. These results indicate
that the proposed continuous-time fusion framework can generalize
well to agile FPV-style trajectories and maintain competitive accuracy
even under few-anchor conditions.

\subsection{Experimental Evaluation in Real-World Scenarios}

\begin{figure}[th]
\centering \includegraphics[scale=0.1]{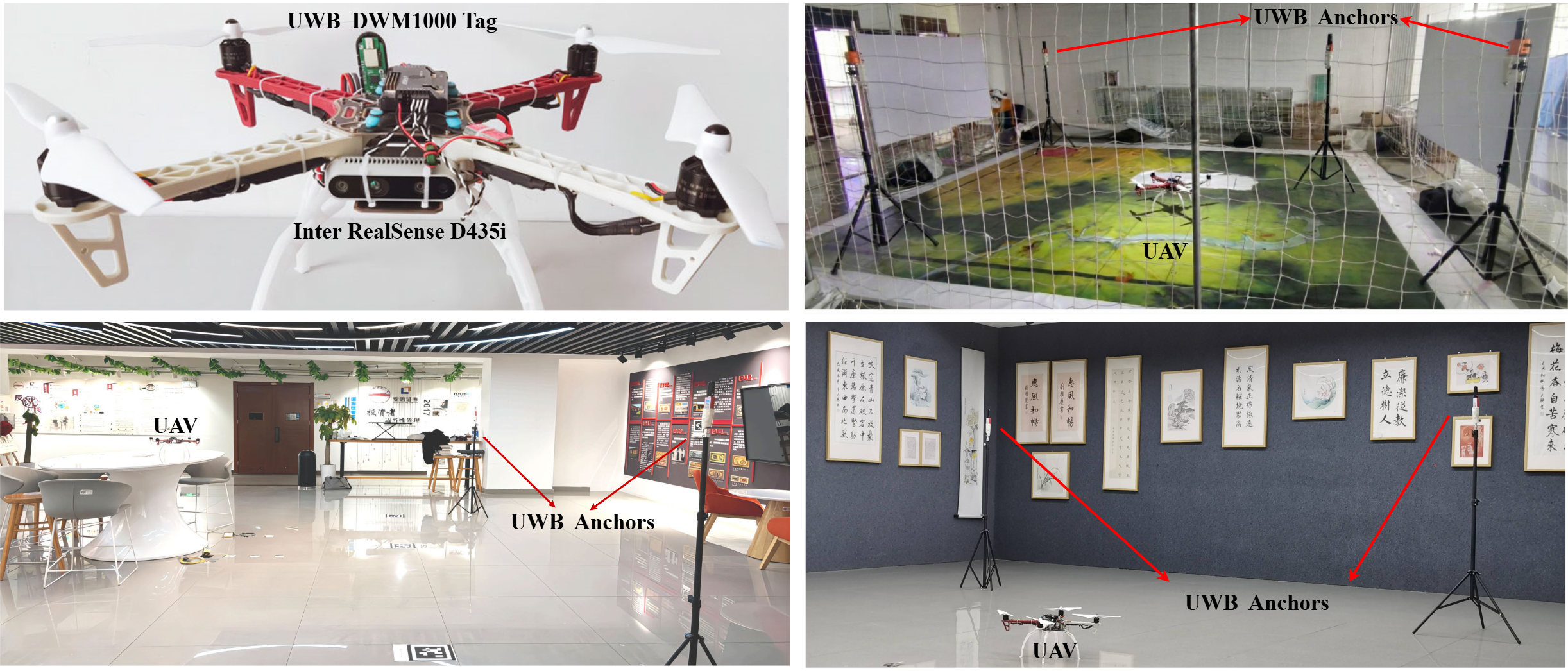}
\protect\caption{Experimental setup and hardware configuration.}
\label{figure:8}
\end{figure}

In this section, to evaluate the performance of the proposed method,
we collect datasets in real-world environments. Specifically, data
are recorded in an underground parking lot, a classroom, and an office
hall. The platform is a small UAV equipped with an Intel RealSense
D435i depth camera and a DWM1000 UWB radio module, as shown in Figure~\ref{figure:8}. Ground-truth trajectories
are obtained using the AprilTag-based method. In addition
to comparing with state-of-the-art methods, we also conduct a set
of ablation experiments under a reduced-anchor configuration.

\begin{table}[!h]
	\caption{Comparison of localization accuracy on real-world scenarios dataset.}
	\label{Tab5}
	\centering
	\footnotesize
	\setlength{\tabcolsep}{3.2pt}
	\begin{tabular}{lccccccc}
		\toprule
		\multirow{2}{*}{Sequence} & \multicolumn{2}{c}{Continuous-time} & \multicolumn{3}{c}{Discrete-time} & \multicolumn{2}{c}{Proposed} \\
		\cmidrule(lr){2-3}\cmidrule(lr){4-6}\cmidrule(lr){7-8}
		& CT-UI & CT-VIO & EKF-VIU & HCCNet & Refloc & Our* & Our \\
		\midrule
		Classroom   & 0.285 & 0.216 & 0.106 & 0.098 & 0.094 & 0.135 & \textbf{0.085}\\
		Office hall & 0.244 & 0.181 & 0.113 & 0.094 & 0.108 & 0.144 & \textbf{0.085}\\
		Underground & 0.335 & 0.297 & 0.135 & 0.109 & 0.115 & 0.134 & \textbf{0.092}\\
		\midrule
		Average     & 0.288 & 0.231 & 0.118 & 0.100 & 0.105 & 0.137 & \textbf{0.087}\\
		\bottomrule
	\end{tabular}
\end{table}

\begin{figure}[!h]
	\centering
	\setlength{\abovecaptionskip}{0pt}%
	\includegraphics[scale=0.18]{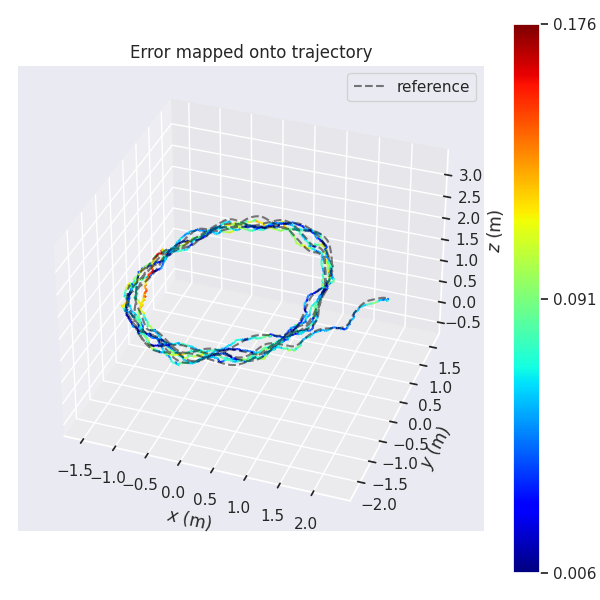} 
\includegraphics[scale=0.18]{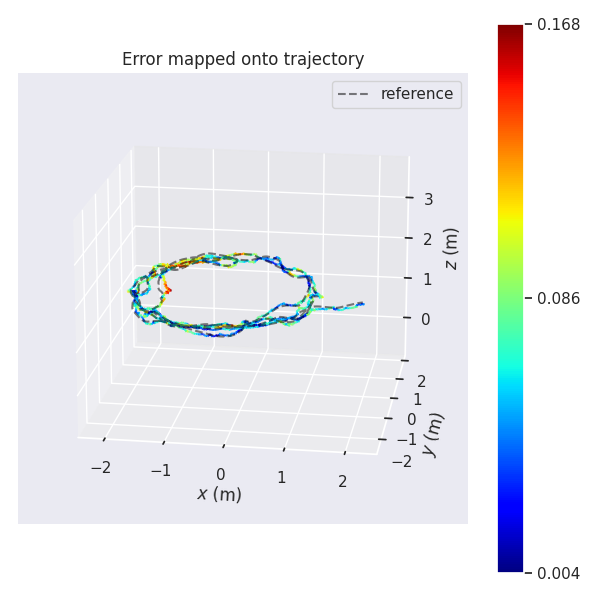}
\includegraphics[scale=0.18]{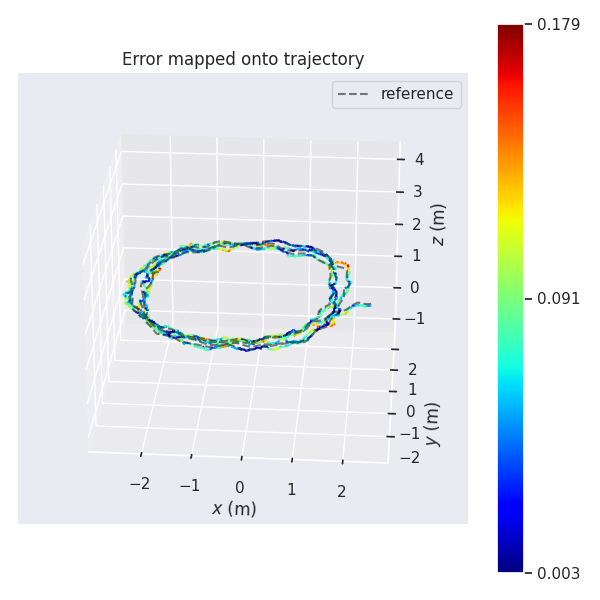}
		
	\caption{Trajectory estimation on the real world sequences.}
	\label{figure:9}
\end{figure}

Table~5 summarizes the localization accuracy on three
real-world scenarios (Classroom, Office hall, Underground). As shown in Figure~\ref{figure:9}, we also visualize the estimated trajectories for qualitative comparison across these scenes. Overall,
the proposed continuous-time VIR fusion with
four anchors (A=4) achieves the best performance on all sequences,
with an average ATE of 0.0873 m. This clearly improves over the continuous-time
baselines Spline-UI and Spline-VIO, showing that
incorporating UWB constraints in the continuous-time spline formulation
substantially enhances accuracy in practical environments. Compared
with the discrete-time VIO-UWB baselines, the proposed method with
A=4 also yields noticeably lower errors than EKF-VIU, HCCNet, and Refloc, particularly in the more challenging
Underground sequence. The three-anchor configuration (A=3) exhibits
a moderate degradation to 0.1374 m on average, and is slightly worse
than the best discrete-time methods, which is consistent with the
reduced geometric constraints under fewer anchors. Nevertheless, A=3
still outperforms the continuous-time baselines, indicating that the
proposed framework remains effective and robust even when the anchor
deployment is not fully adequate.

\section{Conclusion}

This paper proposed a spline-based continuous-time state estimation framework for visual-inertial-ranging fusion localization in anchor-scarce and range-degraded environments. By leveraging VIO motion priors for robust ranging preprocessing and constructing virtual anchors to enhance effective ranging geometry, the method integrates visual, inertial, and (physical/virtual) ranging constraints in a sliding-window B-spline factor graph to obtain a smooth and temporally consistent trajectory estimate. Experiments on public datasets and real-world flights demonstrate improved accuracy and robustness compared with representative continuous-time and discrete-time baselines, especially under reduced-anchor configurations. Future work will focus on tighter uncertainty-aware modeling of virtual anchors and ranging biases, as well as further improving efficiency and real-time deployment on resource-constrained platforms.

 \bibliographystyle{elsarticle-num}
\bibliography{acompat,reference}

\end{document}